\pdfoutput=1 

\documentclass[10pt,twocolumn,letterpaper]{article}

\usepackage{cvpr}              %

\usepackage[table]{xcolor}

\newcommand{\datasetname}{Visual Aptitude Dataset}

\definecolor{cvprblue}{rgb}{0.21,0.49,0.74}
\usepackage[pagebackref,breaklinks,colorlinks,citecolor=cvprblue]{hyperref}
\usepackage{multirow}
\usepackage{color,soul}

\usepackage{amsmath,amsfonts,bm,relsize,dsfont}

\usepackage{color}
\usepackage{xcolor}
\usepackage{wrapfig}
\usepackage[table]{xcolor}
\usepackage{tabularx}
\usepackage{listings}

\lstdefinestyle{base}{
  language=C,
  emptylines=1,
  breaklines=true,
  basicstyle=\ttfamily\color{black},
  moredelim=**[is][\color{red}]{@}{@},
  moredelim=**[is][\color{blue}]{*}{*},
  moredelim=**[is][\color{green}]{~}{~},
}

\newcolumntype{Y}{>{\raggedright\arraybackslash}X}
\newcommand{\myparagraph}[1]{\vspace{0pt} \noindent \textbf{#1}}

\usepackage[printonlyused]{acronym}
\acrodef{LLM}{Large Language Model}
\acrodef{SM}{Supplementary Material}

\usepackage[normalem]{ulem}

\def\paperID{16359} %
\def\confName{CVPR}
\def\confYear{2024}

\title{A Vision Check-up for Language Models}

\author{
Pratyusha Sharma\thanks{Indicates equal contribution.} \quad Tamar Rott Shaham\footnotemark[1] \quad Manel Baradad \quad Stephanie Fu \\ 
 Adrián Rodríguez-Muñoz \quad Shivam Duggal 
\quad Phillip Isola \quad Antonio Torralba
\\
MIT CSAIL\\
}

\begin{document}
\maketitle
\begin{abstract}

What does learning to model relationships between strings teach \acp{LLM} about the visual world? 
We systematically evaluate \acp{LLM}' abilities to generate and recognize an assortment of visual concepts of increasing complexity and then demonstrate how a preliminary visual representation learning system can be trained using models of text. As language models lack the ability to consume or output visual information as pixels, we use code to represent images in our study. Although \ac{LLM}-generated images do not look like natural images, results on image generation and the ability of models to correct these generated images indicate that precise modeling of strings can teach language models about numerous aspects of the visual world. Furthermore, experiments on self-supervised visual representation learning, utilizing images generated with text models, highlight the potential to train vision models capable of making semantic assessments of natural images using just \acp{LLM}.

\let\oldthefootnote=\thefootnote
\renewcommand{\thefootnote}{}
\footnotetext{Project page: \url{https://vision-checkup.github.io/}}
\let\thefootnote=\oldthefootnote

\end{abstract}

\begin{figure*}
    \centering
    \vspace{-1em}
    \includegraphics[width=2\columnwidth]{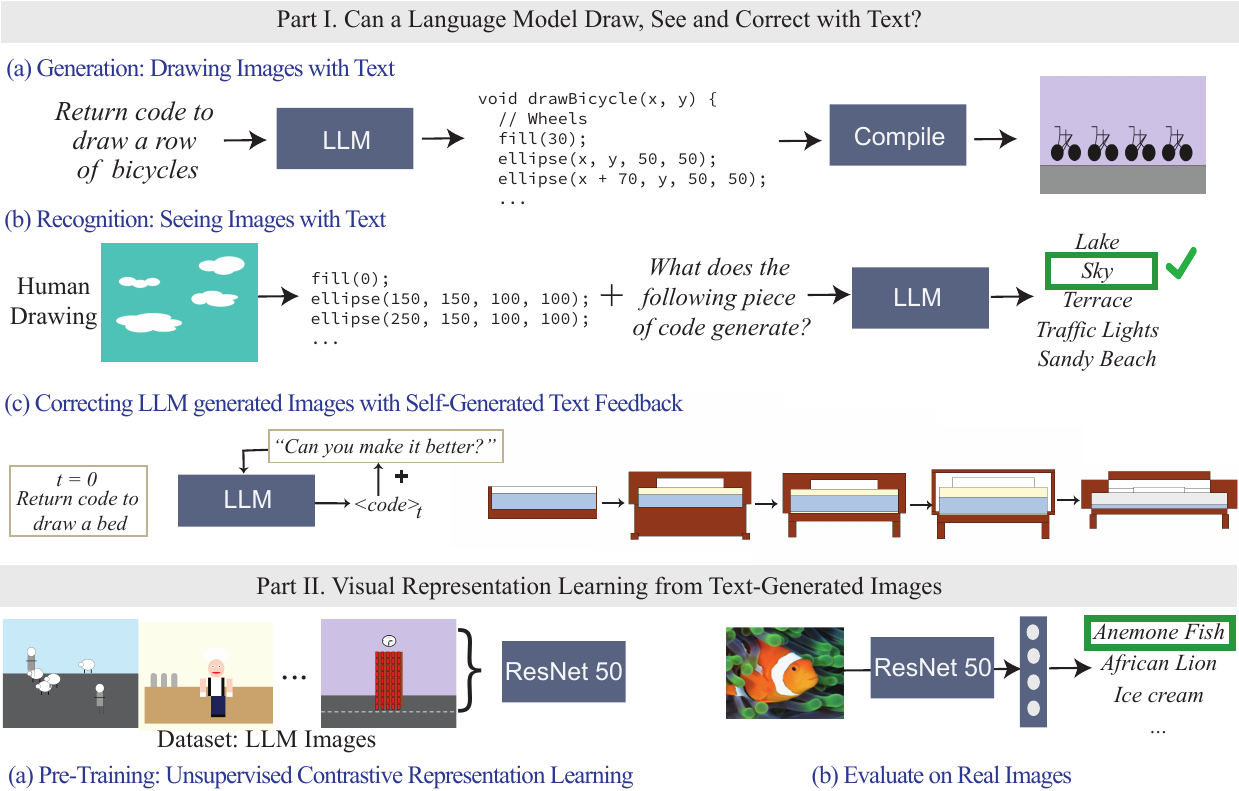}
    \vspace{-0.5em}
    \caption
    {\textbf{Vision check-up for LLMs.} I. Testing the visual knowledge of Language Models. We suggest a set of tests to check the vision abilities of language models, these include (a) the ability to write code that renders complex visual concepts (b) recognizing visual concepts from code (c) correcting rendering code with text-only self-feedback. II. We test whether LLMs can generate data to train a high-performance vision system that can be used to make semantic judgments on natural images. 
    }
    
    \vspace{-1.3em}
    \label{fig:teaser}
\end{figure*}

\section{Introduction}

What does it mean to understand the visual concept of \eg a ``frog''? Is it sufficient to know the color of its skin, the number of feet it has, the location of its eyes, details about how it looks when it hops?  While a subset of this information about the appearance 
of a frog can be acquired from text, it is widely believed that to understand the concept visually, one would need to observe an image of a frog or, better still, observe it from different perspectives and in various real-world scenes. However, to what extent can we learn about the visual ``meaning'' of different concepts from text\footnote{While the question of what can be learned about the visual world from natural language alone is interesting, in our case, ``text'' represents the space of all strings on the internet, including natural language and code.}?

Despite operating on textual data, language model representations have been shown to contain information about named perceptual concepts like shape and color ~\cite{li2021implicit,tenney2019you,chen2020constructing} and have been found to be linearly translatable to representations learned by vision models \cite{merullo2022linearly, tewel2021zero}. These experiments demonstrate that independently, vision and language models represent aspects of the world similarly. 
While investigating model representations for a pre-selected set of attributes can inform us about information encoded by the model, %
it limits studying a fixed set of attributes at once and requires access to the model's internal parameters. 
Instead, as seen in Fig.~\ref{fig:teaser}, we ask: 
\begin{enumerate}
    \item \emph{What do language models know about the visual world?} 
    \item \emph{Can we train a vision system for natural images using a text-only model?}
\end{enumerate}

\begin{figure*}[t]
    \centering
    \includegraphics[width=0.97\linewidth]{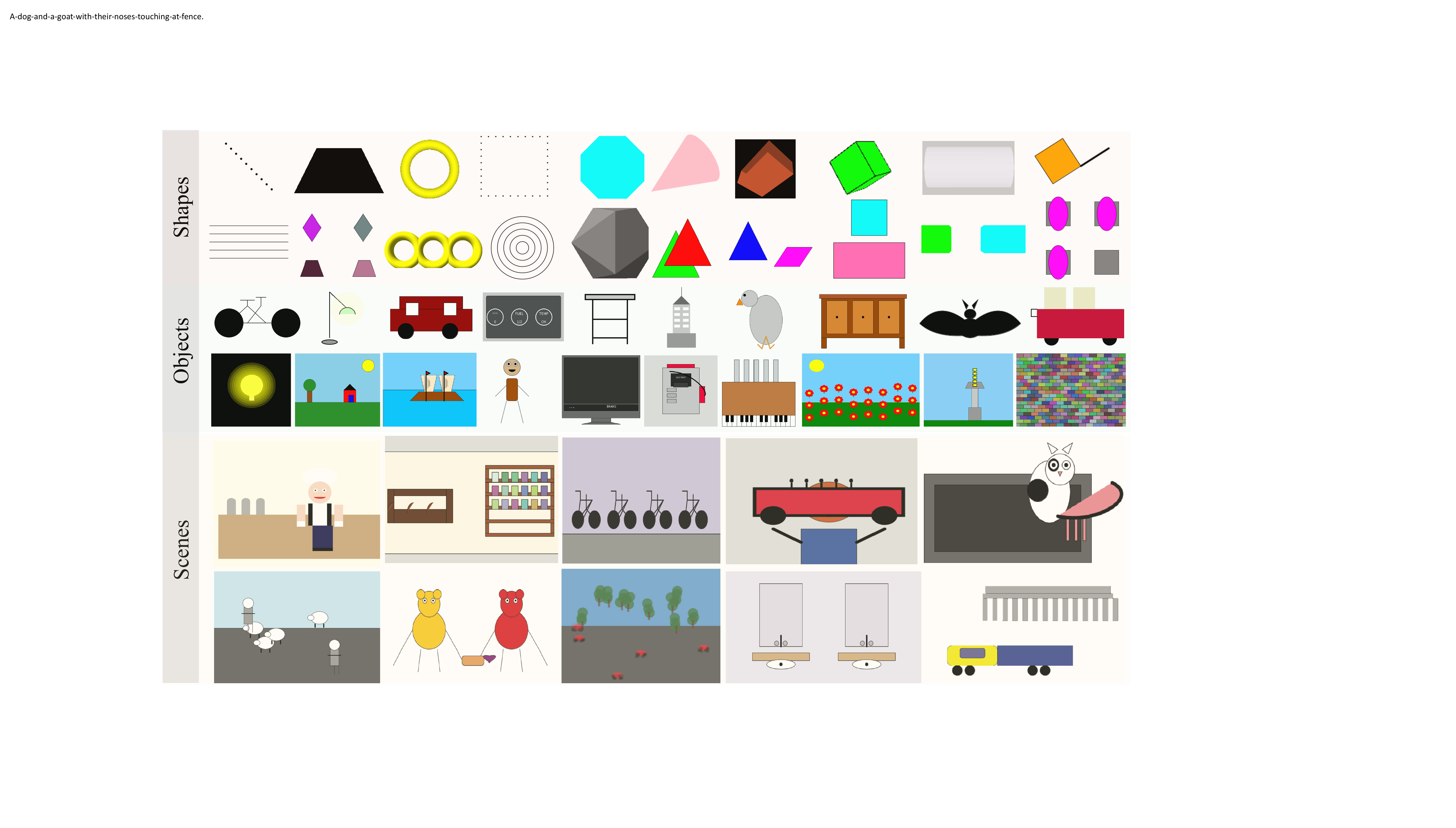}
    \vspace{-2mm}
    \caption{\textbf{Visual Aptitude Dataset.} We collect a dataset of visual concepts of including  shapes, objects and scenes, and ask LLMs to generate corresponding images using a \colorbox{red!5}{Text \(\rightarrow\) Code \(\rightarrow\) Image} generation procedure. Guess the captions of the scenes!\protect \footnotemark 
    }
     \label{fig:generation}
    \vspace{-3mm}
\end{figure*}

To answer these questions, we evaluate what information about the visual world off-the-shelf language models contain by testing their ability to render \emph{(draw)} and recognize \emph{(see)} real-world visual concepts. This allows us to measure their ability to model arbitrary properties, both individually and concurrently, without training a classifier for a rigid set of features one at a time. Although \acp{LLM} are limited in their ability to generate images using pixels, examples presented by ~\cite{bubeck2023sparks} suggest that models like GPT-4 can generate code capable of rendering objects like a unicorn. We take this further by measuring \acp{LLM} abilities to generate visual concepts of increasing complexity via a \textit{textual prompt $\rightarrow$ code $\rightarrow$ image} procedure.
Figure~\ref{fig:generation} shows examples of complex scenes generated by \acp{LLM}. We find that \acp{LLM} are surprisingly good at generating intricate visual scenes composed of multiple objects, effectively modeling spatial relations between them. 
However, there are aspects of the visual world that \acp{LLM} fail to capture, including objects' properties like their textures, precise shapes, as well as surface contact with other objects in the image. 

Next, we evaluate the ability of \acp{LLM} to recognize (see) perceptual concepts (Fig.~\ref{fig:teaser} Part I (b)). We do this by collecting human drawings represented as code and asking \acp{LLM} \emph{what} they depict. The code describes the ordered sequence of shapes, their locations, and colors. We find that unlike humans, where creation is hard and verification is easy, models struggle to interpret/recognize code describing detailed scenes that they themselves can effectively generate. 

Further, we demonstrate that the visual generation competence of a language model can be improved using text-based corrections. We do this by closing the feedback loop between the \acp{LLM} and itself. Here, we first use the language model to generate code illustrating a concept. Following that, the model is repeatedly called by conditioning its generation on its previously generated code and prompted to ``improve its generated code''. We find that making such iterative calls to the model results in improved visual depictions, as shown in Fig.~\ref{fig:teaser}~(Part I (c)).

Finally, we study if \ac{LLM}-generated images could serve as a data source for pre-training vision models and compare them to synthetically generated and natural images. By constructing a  pre-trained visual representation system from \emph{only} text that transfers well to tests on natural images, we demonstrate that text models capture aspects of the visual world similar to those present in natural images. 

To summarize, the paper's main contributions are: 

\begin{enumerate}
    \item \textbf{The \datasetname}: Introducing a hierarchical visual categories dataset consisting of shapes, objects, and scenes descriptions to test the visual capabilities of language models.
    \item \textbf{Generation}: Testing and quantifying the generation capabilities of LLM's, showing that it is possible to generate detailed and diverse scenes using text-only models. We also show that it is possible to improve the quality of the generated images by using text-based feedback.
    \item  \textbf{Recognition:} Analyzing whether \acp{LLM} are also able to recognize image generative code as well as producing it. We test this capability using out-of-distribution samples generated by humans, which we crowdsource.
    \item \textbf{Training for natural image tasks without natural images}:
    We show that the images generated by LLMs are useful for training visual backbones, achieving state-of-the-art performance when complemented with other procedurally generated image datasets.
\end{enumerate}

\section{Related work}
\myparagraph{Vision and language models:}
Language models have been shown to be extremely powerful in understanding and generating visual information when paired with vision models~\cite{wu2023visual}, training vision adaptors~\cite{eichenberg2021magma,zhu2023minigpt,liu2023visual,merullo2022linearly}, 
or when trained jointly over visual and textual data~\cite{tewel2021zero,tsimpoukelli2021multimodal,yang2023mm}. Although vision-language pre-training / chaining vision and language models allow models to reason about aspects of the visual world, we investigate the visual capabilities of models representing images with text. Furthermore, several benchmarks have been proposed to evaluate the ability of LLMs on textual tasks~\cite{gu2022language,hong2021ptr,srivastava2022beyond,mahowald2023dissociating,hernandez2017evaluation}. 
Unlike them, we propose a procedure to evaluate LLM's \emph{vision} abilities.

\myparagraph{Visual Understanding in Language Models.} Meaning Representations ~\cite{li2021implicit,tenney2019you,chen2020constructing}
show that language models contain information of the state of the world in their internal representations that can be recovered by probing these models for attributes of interest. Additionally, ~\cite{patel2022mapping, Abdou2021CanLM, wei2022emergent} demonstrate that language models are capable of representing visual concepts such as ``color'' and ``shape''. 
However, the attributes studied have been limited to only those that can be 
described by natural language and can only be investigated one at a time. 
Moreover, with models being placed behind closed walls, it becomes increasingly difficult to probe the internal representation of models for the presence of visual attributes of interest. 

\begin{figure}[t]
    \centering
    \includegraphics[width=0.98\linewidth]{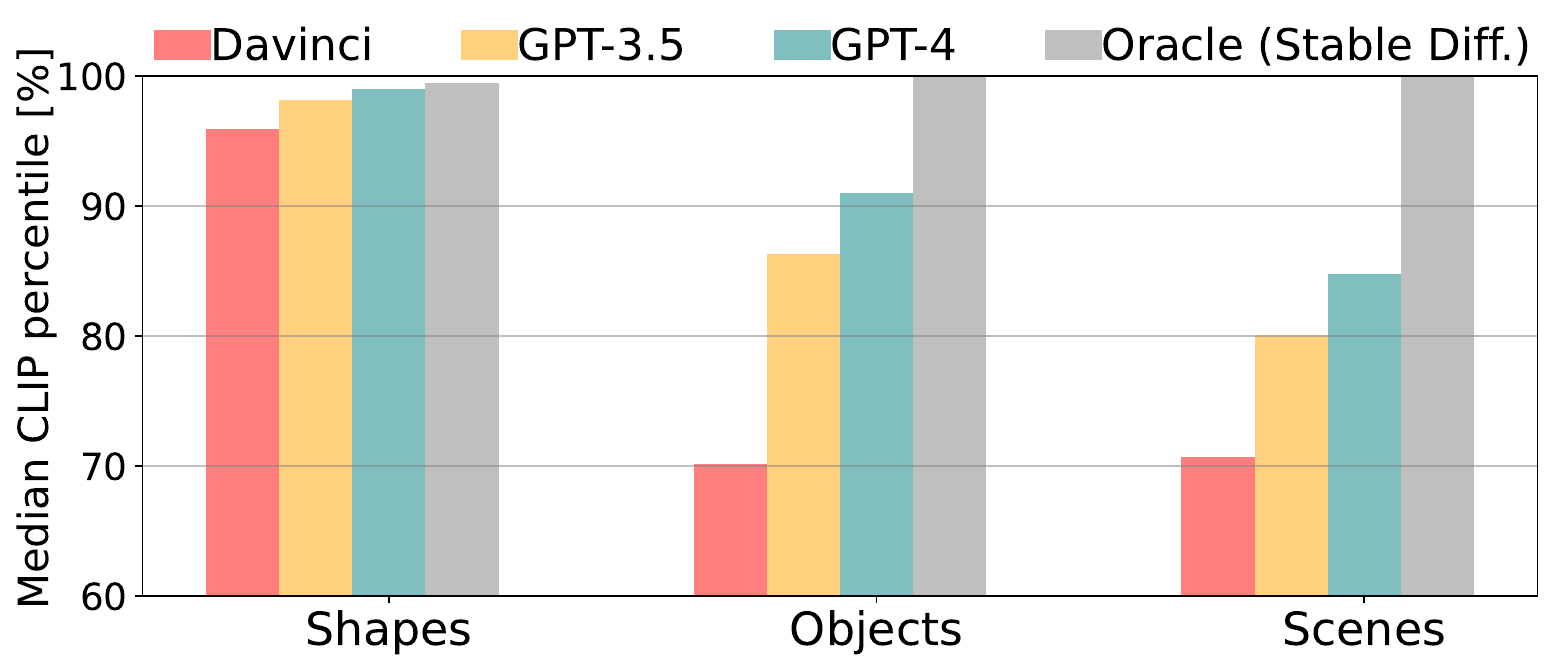}
    \caption{\textbf{Image-Text Fidelity.} Median CLIP image-text retrieval percentiles of images generated by different \acp{LLM}. We include Stable Diffusion as an Oracle. Chance is $50\%$.}\label{fig:clip}
    \vspace{-1.2em}
\end{figure}

\myparagraph{Program synthesis via LLMs.} Pioneered by OpenAI's Codex \cite{Chen2021EvaluatingLL}, Github's Copilot \cite{GithubCopilot}, Meta's Code Llama \cite{CodeLlama} etc., LLMs have been shown to possess exceptional coding capabilities \cite{zhang2023survey}. 
Recently, Bubeck \etal \cite{bubeck2023sparks} highlighted the emergent properties of a GPT-4 model for image generation / scalable vector graphics via text-prompted TikZ or javascript codes. In this work, we build on their insights and carefully examine the diversity and realism of multiple text-only language models like GPT-3.5, Davicini, GPT-4 (see Fig.~\ref{fig:clip}, Fig.~\ref{fig:diversity}) for programmable image generations. Furthermore, as one of our key contributions, we analyze the usefulness of the procedurally generated images for self-supervised visual representation learning (see Sec.~\ref{vision-ssl}).

\myparagraph{Training vision models with synthetic data.}
The ability to train a vision system using synthetic data was studied in several tasks including optical flow~\cite{barron1994performance,mayer2016large,dosovitskiy2015flownet}, segmentation~\cite{chen2019learning,ros2016synthia}, detection~\cite{rozantsev2015rendering}, classification~\cite{azizi2023synthetic,sariyildiz2023fake}, and representation learning~\cite{tian2023stablerep}. Perhaps the most related set of work
studied how to train vision backbones using images generated from human-written code which capture different priors of the visual world, such as textures and shapes. These include generative processes like fractals \cite{connor_fractals,fractals}, dead leaves \cite{baradad2021learning}, sinusoidal waves \cite{takashima2023visual}, and even a crowd-sourced dataset of thousands of generative programs \cite{baradad2022procedural}. While these approaches achieve competitive performance, it remains unclear how to systematically introduce high-level semantic concepts related to shapes and scene composition without the intervention of human experts.

\footnotetext{\textbf{Captions for Fig~\ref{fig:generation} scenes:} (left to right, top to bottom) \footnotesize \textbf{C}hef standing next to a counter with jars; \textbf{O}ffice with leather couch, surrounded by books; \textbf{R}ow of bicycles; \textbf{B}irthday boy with car shape cake \& candles; \textbf{B}lack \& white cat sitting on side of a computer monitor; \textbf{C}ouple of men hearding sheep down the road;  \textbf{T}wo stuffed animals cutting bread \& spreading jelly on it; \textbf{B}lurred image of motorized scooters on wooded road. \textbf{B}athroom with two sinks \& tow mirrors. \textbf{Y}ellow \& blue train is next to an overhang;}

\section{\datasetname: Points to Scenes}
We evaluate an \ac{LLM}'s visual competence by measuring its ability to create, recognize, and modify image-rendering code on a hierarchy of concepts. This resulting dataset of images also serves as the corpus used for pre-training a vision model in the later part of the paper. 
We construct three datasets with textual descriptions of visual concepts of gradually growing complexity. Starting with simple shapes and their compositions, to objects, and finally to complex scenes described by elaborate natural language descriptions. 
Samples from the dataset can be found in Fig.~\ref{fig:generation} and in the \ac{SM}. 

(i) \emph{Shapes and their compositions:} The first dataset contains a composition of shapes from different categories such as points, lines, 2D-shapes, and 3D-shapes with 32 different attributes like color, texture, location, and spatial arrangement. The full dataset contains more than 400K examples, and we sample 1500 for tests in our experiments.

(ii) \emph{Objects:} The second dataset contains the 1K most frequent objects of the ADE20K dataset~\cite{zhou2017scene,zhou2019semantic}. Objects are more difficult to generate and recognize than shapes, as they contain complex compositions of many shapes.

(iii) \emph{Scenes:} The last dataset consists of complex scene captions describing diverse places with multiple objects. For this, we uniformly sample 1000 scene descriptions at random from the MS-COCO~\cite{lin2014microsoft} dataset.

\subsection{Representing images with code} In the dataset, the visual concepts are described with language. 
For instance, we can describe a scene as ``a sunny summer day on a beach, with a blue sky and calm ocean.'' 
We test the visual competence of \acp{LLM} by prompting them with these descriptions and measuring their ability to generate code that can be compiled to render images depicting the scenes. Why code?
While \acp{LLM} can sequentially output pixel values to generate images \cite{bubeck2023sparks} their ability to do so is currently limited. %
Code, on the other hand, can provide a descriptive yet concise representation of the visual world. It can be used to represent higher-level perceptual attributes and language models are already trained on examples of code. In this paper, code in different programming languages will serve as the primary mode of interaction and visual scene generation for \acp{LLM}.

\begin{figure}[t]
        \centering
        \includegraphics[width=0.98\linewidth]{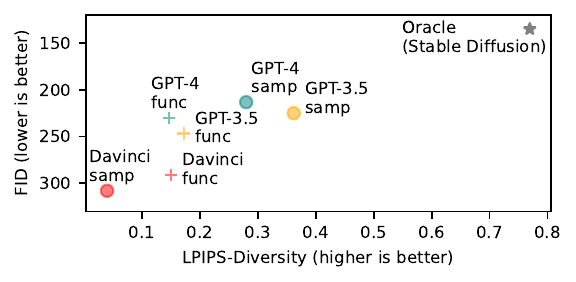}
        \vspace{-0.5em}
        \caption{\textbf{Realism vs. Diversity.} With both sampling strategies, LLMs are able to draw diverse illustrations of the same concept.} 
        \vspace{-1.2em}
        \label{fig:fid_div}
\end{figure}

\subsection{Models and Programming Languages tested} For this study, we evaluate four language models, each tested on four different programming languages. 

(i) \emph{Language models:}  
We evaluate the visual competence of GPT-3.5 (both \texttt{text-davinci-003} and \texttt{GPT-3.5-turbo} models) and GPT-4\footnote{There are two types of GPT-4. We interact here with GPT-4 model and not the GPT-4(V) model.}. Models, like Llama2 (chat 70B), GPT-J, and GPT-2 failed to generate executable image-rendering code reliably and are excluded from analysis in the main paper but are included in the SM. 

(ii) \emph{Programming languages.} To validate that a model's visual capability is not restricted to a specific programming language we use four programming languages with different expressiveness. These are: python-matplotlib, python-turtle, Processing (built over Java), and TikZ (built over Tex). A model's ability to generate and recognize the same visual concept across programming languages %
hints at the model possibly having a coherent and language-agnostic representation of that concept.

\section{A Vision Checkup for \acp{LLM}}

In this section, we evaluate the visual capabilities of \acp{LLM}. The models are evaluated on three tasks: (i) Generation / Drawing with text: assesses an \ac{LLM}'s competence in generating image-rendering code corresponding to a specific concept. (ii) Recognition / Seeing with text: tests the \acp{LLM}'s performance in recognizing visual concepts and scenes represented as code. We test each model on the code representation of human drawings. %
(iii) Correcting drawings with text feedback: evaluates an \ac{LLM}'s ability to iteratively modify its generated code using natural language feedback generated by itself.

\subsection{Generation: Drawing with Text}
\label{subsec:generation}
\myparagraph{Experimental Setup.}
To test what \acp{LLM} are capable of visualizing, we evaluate their ability to generate code representing concepts from our Visual Hierarchy dataset across four programming tools. The prompt fed to the \ac{LLM} is:

\hl{Prompt: \emph{``write code in the programming language "[programming language name]" that draws a [concept]''}}. 

We then render the images by compiling the code in its corresponding programming language. %
Additional details about the experiment protocol can be found in the \ac{SM}.
\smallbreak

\begin{figure}[t]
    \centering
    \includegraphics[width=1.02\linewidth]{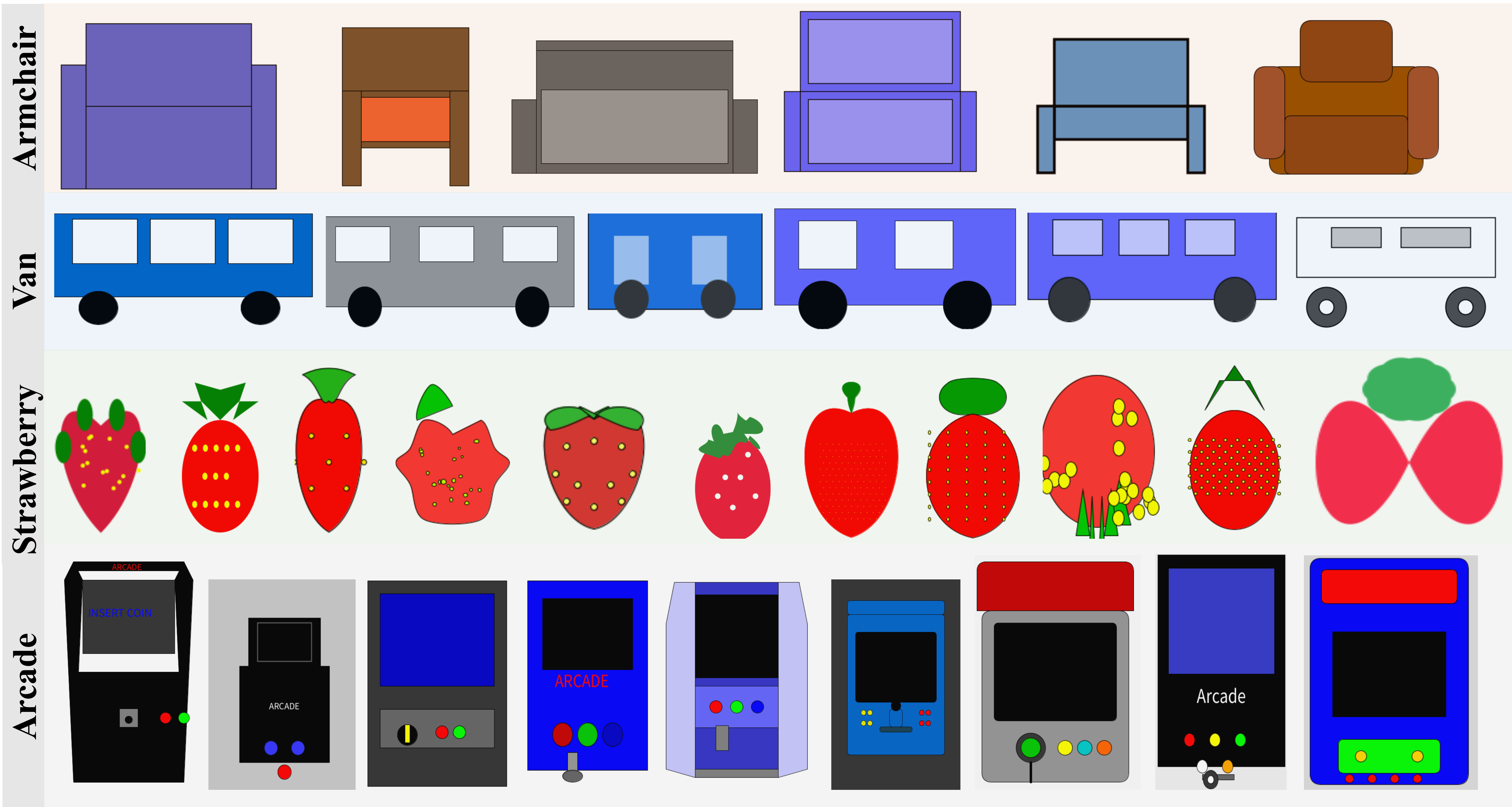}
    \vspace{-4mm}
    \caption{\textbf{Diversity.} LLMs are capable of generating diverse meaningful instances of the same concept, showcasing their ability to represent concepts beyond a single fixed prototype.}  
     \label{fig:diversity}
    \vspace{-1.2em}
\end{figure}

\myparagraph{Evaluation protocol.}
We evaluate the visual quality and diversity of the images rendered using the following metrics. 

(1) \emph{Fidelity}: %
We compute the fidelity between the generated image and its ground-truth caption by 
retrieving the best caption for the image. %
We first calculate the agreement between each image and all potential captions within the same category (shapes/objects/scenes) using their CLIP score. We then report the rank of the ground-truth caption in percentage (\eg a score of $100\%$ implies that the ground-truth concept is ranked first). %
In the \ac{SM}, we augment this measure with human perceptual studies and show that the CLIP rank reflects human preference as evidenced from their high correlation ($r = 0.82$,p-val$=1.25e^{-09}$).

(2) \emph{Diversity:} To assess the ability of the models to render diverse content, we use the LPIPS-diversity score~\cite{zhang2018unreasonable} over image pairs representing the same visual concept. 

(3) \emph{Realism:} For a uniformly sampled collection of 1K images from ImageNet~\cite{imagenet_cvpr09}, we use the Fr\'echet Inception Distance (FID)~\cite{heusel2017gans} to quantify the difference in the distribution of natural images and those generated by \acp{LLM}. 

\underline{Baseline:} As an oracle, we include images generated by a text-to-image model (Stable Diffusion~\cite{rombach2022high}) and report their scores across all evaluation metrics. \smallbreak

\smallbreak
\myparagraph{What can \acp{LLM} visualize?} 
We find that \acp{LLM} can visualize real-world concepts from across the visual hierarchy. Examples of LLM-rendered images can be found in Fig.~\ref{fig:teaser},~\ref{fig:generation},~\ref{fig:diversity}, and in the \ac{SM}. \acp{LLM} are capable of generating non-trivial visual compositions, examples of such are shown in Fig.~\ref{fig:generation}; The model composes two unrelated concepts (``car shaped cake''), generates visual phenomena (``blurred image''), and manages to correctly interpret spatial relations (\eg``a row of bicycles'' arranged horizontally.) 
Unsurprisingly, the competence of the models deteriorates with increasing concept complexity from shapes to scenes, as seen in the median image-text retrieval CLIP-scores across the different categories Fig.~\ref{fig:clip}. For more complex visual concepts such as drawing scenes comprising multiple objects, GPT-3.5 and GPT-4 are more accurate at drawing scenes with intricate descriptions using processing and tikz than python-matplotlib and python-turtle. 
For objects and scenes, CLIP scores indicate that concepts containing "person", "vehicle", and "outdoor scenes" are the easiest to draw (see full analysis in the \ac{SM}). 
This ability to render complex scenes comes from the expressivity of the rendering code, the model's programming capability in each of them, and the quality of its internal representations of the different concepts involved.

\smallbreak
\myparagraph{What can \acp{LLM} not visualize?}
In some cases, even relatively simple concepts are difficult for the models to draw. We identify several common failure modes: (a) Language models specifically struggle with concepts combining a set of shapes and a specific spatial organization, (b) Drawings are coarse and lack detail. These are the common failure cases for Davinci, especially when coding with matplotlib and turtle. (c) Depiction is incomplete, corrupted, or represents only a subset of the concepts (typical for the scenes category). 
An interesting standalone failure case is drawing digits. With all models and languages, we found this task to be challenging. 
See \ac{SM} for a discussion of failure cases and the effect of prompting on the model's generation. \smallbreak

\myparagraph{Diversity and Realism.}
Language models exhibit the ability to generate diverse visualizations of the same concept as seen in Fig.~\ref{fig:diversity}. To generate different samples of the same scenes, we compare two strategies: (i) Repeated sampling from the model (ii) Sampling a parametrized \emph{function} that allows creating a new drawing of the concept by changing parameters. The ability of the model to render diverse realization of visual concepts is reflected in the high LPIPS-diversity scores in Fig.~\ref{fig:fid_div}. 
The ability to generate diverse images suggests that \ac{LLM}s can represent the visual concept in many ways rather than a limited set of prototypes. \ac{LLM} generated images are far from as realistic as natural images, with the models scoring poorly on the FID metric as compared to the Stable Diffusion oracle (Fig~\ref{fig:fid_div}).
However, it is interesting to note that modern models rank better than older models, indicating that they might be slowly inching towards increasingly realistic representations of the world.

\subsection{Recognition: Seeing with Text} 

\begin{table}
\centering
\begin{tabular}{@{}clrrr@{}}
\toprule
 Model Name & Objects & Scenes \\ \midrule

Davinci  & 0.136 & 0.221\\
 GPT3.5 & 0.228 & \textbf{0.380}\\
 GPT4 &  \textbf{0.234} & 0.212 \\
\midrule
Baseline [Chance] & 0.2 & 0.2\\
\bottomrule 
\end{tabular}

\caption{\textbf{Recognition of Human Drawings.} Models struggle to correctly classify human drawings into their categories. While GPT-3.5 correctly classifies images over chance for the scenes category, other models classify images correctly barely over chance.}\label{tab:rec}
\vspace{-0.2in}
\end{table}

\begin{figure*}[h]
    \centering
    \vspace{-1.2em}
    \includegraphics[width=0.98\linewidth]{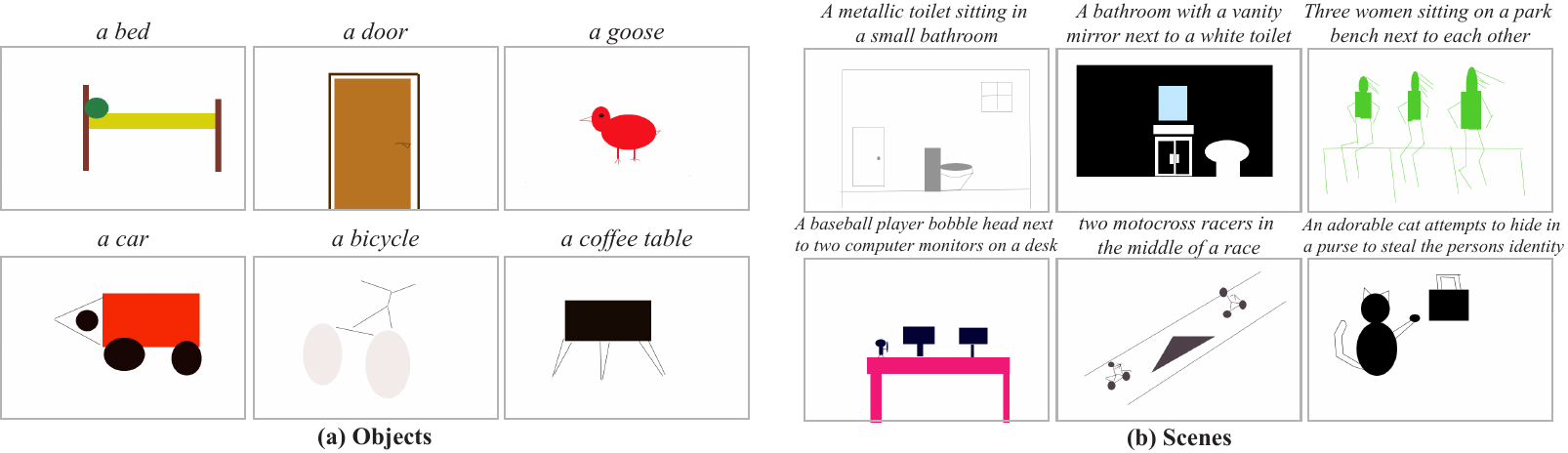}
    \vspace{-0.8em}
    \caption{\textbf{Human drawings.} Examples of drawings made by users using our drawing interface that passed the CLIP score filtering. Each of the collected drawings is converted into processing code and is then included in the \ac{LLM}'s recognition tests. 
    }
    \label{fig:human_drawings}
\end{figure*}

Recognizing the contents of an image requires inferring how elements such as points, strokes, colors, and shapes combine spatially to give rise to objects that are themselves composed to represent complex scenes. 
Evaluating a model's ability to recognize image-rendering code offers insights into its competence in interpreting high-dimensional representations of visual scenes, including ones very different from its ``memorized'' prototypes. 
While code on the internet can be a good source to test these models, it contains comments, print statements, and symbol names that reveal essential semantic information about what the code represents and may not actually require sophisticated inference over the steps used to create the drawing. 
Further, we wanted to ensure that the code used for evaluation recognition was new and not part of the training set for any of the models tested. 
Therefore, we collect our own dataset of code representing drawings as seen in Fig.~\ref{fig:human_drawings}.\smallbreak

\begin{figure*}
    \centering
    \includegraphics[width=2.1\columnwidth]{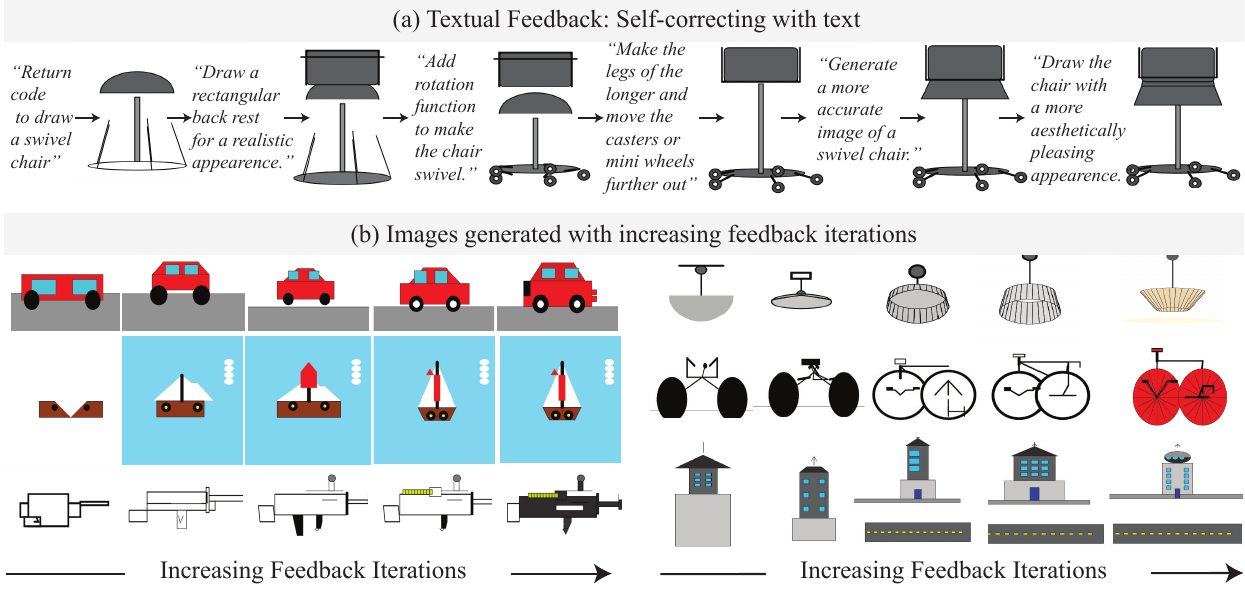}
    \vspace{-1.2em}
    \caption{\textbf{Improved visual generation with text feedback.} The improvement in the visual generation of models due to feedback is oftentimes gradual, with the addition of a few features at a time over the course of the feedback process. Conditioning on the code generated from the previous iterations, to a limited degree, ensures that the model is constrained to modifying aspects of the current image. 
    }
    \label{fig:feedback}
    \vspace{-5mm}
\end{figure*}

\myparagraph{Dataset.} 
While most people excel at drawing images using ``Paint''-like applications, writing code that represents images is trickier and requires more skill.
To test the recognition ability of models we collect a dataset of human drawings and their corresponding representation in code. This dataset was constructed following a setting similar to the game Pictionary or Google's ``Quick Draw!'' application. Users were prompted to draw a concept from the Visual Aptitude Dataset within a two-minute time limit using a drawing interface. The interface recorded their drawing steps in the Processing programming language. Shapes were excluded from the analysis, and the generated images were filtered for quality using the Fidelity score.
The interface was designed to mimic the components expressed in drawings made with processing and to encourage users to portray the entire prompt rather than a detailed but incomplete illustration in the allotted time. We collect 600 images per category, of which 162 object images and 113 scene images pass the fidelity check (rank$\leq 40$) and are used for evaluation. For further details, refer to the \ac{SM}.\smallbreak

\myparagraph{Evaluation protocols.} We evaluate the models' capability of recognizing visual concepts represented as code by measuring the model's ability to classify the image in a 
\emph{multi-class classification} setting. In this test, the model is prompted with the code and a list of visual concepts, where one of the list's entries describes the code accurately. The model is then tasked with matching the visual concept to the correct entry. The prompt to the model is:

\hl{Prompt: \emph{"Which of the following concepts from the list does the [code] represent? [concept1, concept2,...]"}.} 

\underline{Baseline:} The success of the chance baseline is decided by the probability of the desired outcome. This is given by $1/N$ where $N$ is the number of outcomes. For the multi-class classification setting with five labels this is 0.2. \smallbreak

\myparagraph{Analysis.} Human drawings present a unique recognition challenge as there is a lot of diversity in the images representing a given concept.

\smallbreak
\myparagraph{Language models can do very limited spatial reasoning.} Table.~\ref{tab:rec} shows that GPT-3.5 beats the chance baseline across the scenes setting, demonstrating that the visual recognition capability of the models is non-trivial, allowing it to (limitedly) recognize code representing human drawings. While the exact mechanism that allows models to do so is unclear, the task requires models to identify objects, their attributes and spatial arrangements. Wrongly attributing any information can result in a completely different visual scene. Although models cannot perfectly recognize all images, the exhibited recognition capability is non-trivial. 

\smallbreak
\myparagraph{Models can fail to recognize concepts they can otherwise draw very well.} Unlike humans, where the ability to draw something well automatically implies the ability to recognize the same concept well, models can fail to recognize concepts they have no problem rendering. This contradicts the notion that creation is hard, but verification can be easy. Failures in recognition, as shown in the \ac{SM}, showcase that images that models fail to recognize are the non-stereotypical depictions of the visual concept, showcasing that there might be limits to a model's ability to recognize perceptual concepts from code.

\subsection{Textual-Feedback: Correcting with Text}

The model's ability to generate an image is limited in part by the prompt. Failure to generate an image corresponding to a particular concept does not necessarily imply its lack of "knowledge" of that particular concept but rather its lack of immediate accessibility. Could direct systematic guidance and textual feedback help improve a model's visual capability? And if so, can this be automated? \smallbreak

\myparagraph{Experimental Protocol.} The visual competence of a generator language model can be improved by pairing it with itself. This procedure serves as prompt-facilitated "self-talk" and helps scope out the model's internal representation of a particular visual concept over multiple rounds of generation. 
Additionally, selective step-wise correction of a model's generation gives us insights into whether the model has memorized a fixed  "prototype" corresponding to a visual concept or if its representation can be systematically modified by iteratively prompting it to improve its drawing.  

\hl{Prompt: \emph{"The following [code] does not accurately represent the [concept]. How can you do better?"}.} \smallbreak

\myparagraph{Evaluation Protocol.}
To evaluate the improvement in the model's drawing capability, we use the fidelity score that was described in Section~\ref{subsec:generation}.

\begin{figure}
    \centering
    \includegraphics[width=0.4\textwidth]{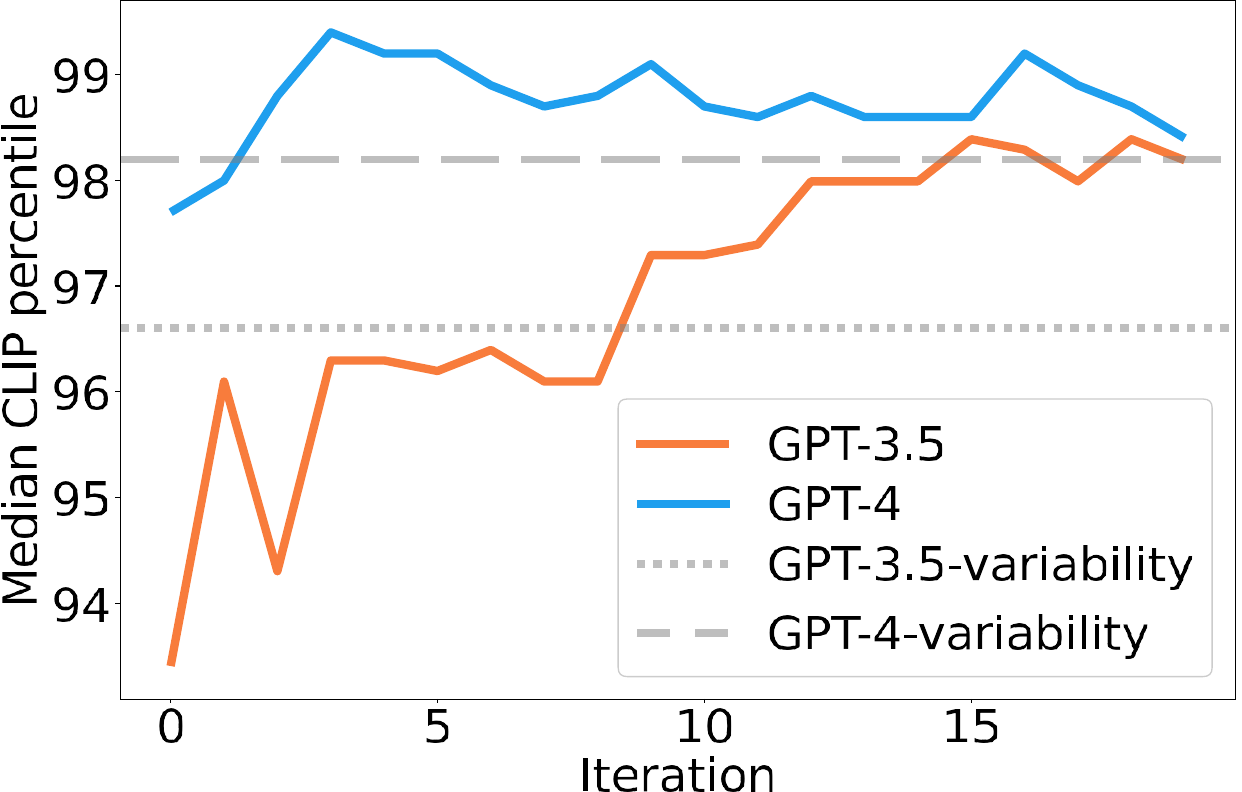}
    \caption{\textbf{Text-feedback improves visual generation competence:} Improvement in the quality of the generations is depicted by increasing median CLIP percentile as a function of feedback iterations. The model with textual corrections outperforms selecting the best image from multiple random samples of the same concept.}
    \label{fig:feedback-clip}  
    \vspace{-0.15in}
\end{figure}

\underline{Baseline:} To assess whether the model's drawing improvements stem from textual feedback or repeated model calls, we conduct a 'variability' baseline. We generate 20 independent outputs for the same concept and report the median CLIP percentile for the best images per concept. Mean variability score is reported in Fig.~\ref{fig:feedback-clip}. \smallbreak
\smallbreak

\myparagraph{The visual competence is improved solely by text-based feedback.}
We present examples of images generated with iterative feedback in Fig.~\ref{fig:feedback}. Visual quality and fidelity 
to the visual concepts significantly improve with multiple rounds of feedback. Fig.~\ref{fig:feedback-clip} shows that GPT-3.5, a weaker generator than GPT-4, can be made to be nearly as competent of an image generator as GPT-4 with text feedback from itself. Therefore, models can benefit in returning an overall better answer from multiple rounds of self-verification. In fact, with 20 rounds of feedback, the average performance of GPT-3.5 approaches that of GPT-4.
Further analysis and visualization for the dataset are in the SM.

\section{Learning a Vision System from Text}
\label{vision-ssl}

Lastly, we test whether \ac{LLM}-generated images can be used to train a generally capable vision system for \textit{natural} images. This relates to abundant existing work \cite{fractals,baradad2021learning,connor_fractals,baradad2022procedural,takashima2023visual} that studies the same phenomena by pretraining vision systems using only procedurally generated images.
\smallbreak

\begin{figure}
     \centering
     \includegraphics[width=0.5\textwidth]{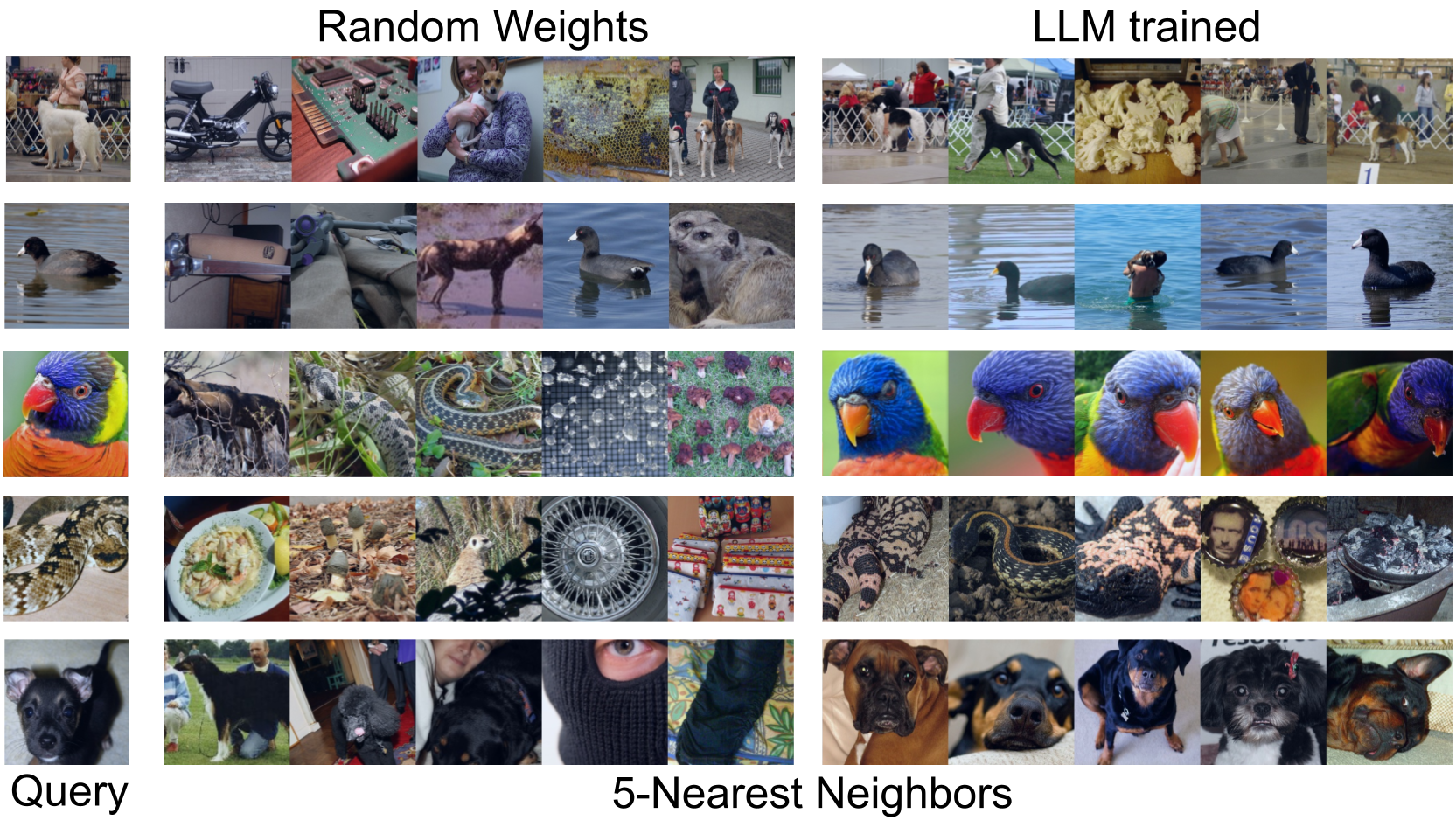}
     \caption{\textbf{Nearest Neighbors Retrieval on ImageNet-100.} Nearest neighbors on ImageNet-100 for a randomly initialized network and a network trained with all our LLM-generated images.}
     \label{fig:nearest_neibhgors_LLM}
\end{figure}

\myparagraph{Training and evaluation protocol.} We use unsupervised contrastive learning, which allows pretraining vision backbones with an unlabeled set of images, and follow the training and evaluation protocol of~\cite{baradad2021learning}. We train a {ResNet-50}~\cite{resnet} using the MoCo-v2 method~\cite{moco_v2} %
over a dataset of 
1.3M $384\times384$ images generated by LLMs, for 200 epochs with batch size of 256. For comparison, we generate datasets of the same size and image resolution with 4 different procedural generation methods~\cite{baradad2021learning,fractals,baradad2022procedural}. After training, 
we evaluate the performance of models trained on each of the datasets using two approaches: (i) training a linear layer on top of the frozen backbone for ImageNet-1k classification for 100 epochs, and (ii) using 5-nearest neighbor retrieval on {ImageNet-100}~\cite{cmc}. The latter is of special interest, as nearest-neighbor retrieval shows that 
models trained solely on LLM %
generated data %
yield powerful representations for natural images, without the need of training a linear layer. This can be qualitatively seen in Fig.~\ref{fig:nearest_neibhgors_LLM}. 
\smallbreak
\myparagraph{Baselines and datasets.} 
We compare our \ac{LLM} generated images against existing procedurally generated images. These include simple generative programs like dead-levaves~\cite{baradad2021learning}, fractals~\cite{fractals}, and StyleGAN~\cite{baradad2021learning}, each consisting of a single program that generates highly diverse images. We also consider the Shaders-21k dataset~\cite{baradad2022procedural}, a large collection of procedural image programs, each one producing a family of images of high diversity. 
Our LLM-generated dataset consists of all the available images obtained using the different LLMs as described in Sec.~\ref{subsec:generation}, a total of 80k images. 
We augment these by randomly sampling convex combinations of six data points using MixUP \cite{mixup} (shown to be effective for other synthetic images~\cite{baradad2022procedural}), and reach a total of 1.3M images.
As an ablation, we report an additional experiment that excludes GPT-4 images from the training (a 31k subset of the total 80k). Additional dataset breakdowns (\ie each LLM individually) are reported in the SM.
\smallbreak

\setlength{\extrarowheight}{1pt}
\begin{table}[]
\centering
\resizebox{\columnwidth}{!}{
\begin{tabular}{llll}
\specialrule{.2em}{.1em}{.1em}
& \multirow{2}{*}{Pre-training Dataset} &\multicolumn{1}{c}{I-1k}  &\multicolumn{1}{c}{I-100} \\
& & Linear  & 5-NN     \\
\specialrule{.1em}{.05em}{.05em}
\multirow{-1}{*}{Random Init.} & None & \ \ 4.36    &  \ \  4.28    \\

\multirow{-1}{*}{Real} & Places              & 55.59    &   57.04    \\
\specialrule{.1em}{.05em}{.05em}
& Dead-leaves         & 20.00     & 12.76        \\ 
& FractalDB-1k        & 23.86      & 17.24 \\ 
& StyleGAN O          & 38.12     &     33.00    \\
\multirow{1}{*}{Procedural}  & S-21k         & 44.83  &  43.24  \\ 
\cmidrule{2-4}

& LLMs (w/o GPT-4)         &  33.60 &   22.42        \\ 
& LLMs (w/ GPT-4)          & 36.16  & 27.44          \\ 
& LLMs (w/o GPT-4) + S-21k & 45.79  & \textbf{43.40}  \\ 
& LLMs (w/ GPT-4) + S-21k & \textbf{46.03}      & 43.36  \\ 
\specialrule{.1em}{.05em}{.05em}
\end{tabular}
}
\caption{\textbf{Learning a vision system.} Top-1 Linear evaluation on ImageNet-1k and 5-NN on ImageNet-100, for a ResNet-50 pretrained with different real and procedurally generated datasets including LLM's generated images.%
}
\vspace{-1.2em}
\label{tab:moco_performance}
\end{table}

\myparagraph{Analysis.}~Table \ref{tab:moco_performance} shows that models trained with only \ac{LLM}-generated images outperform simple datasets like dead-leaves or fractals, but are still inferior to alternatives. 
Through visual inspection of the data, we attribute this inferiority to the lack of texture in most \ac{LLM}-generated images, as observed in the figures throughout the paper and in the SM. 
To overcome this, we combine the Shaders-21k dataset \cite{baradad2022procedural}, which generates texture-rich images with the samples obtained from \acp{LLM}. This is done by sampling with $50\%$ chance from either dataset and applying MixUP. As seen in Table \ref{tab:moco_performance}, these models outperform all procedural generation-based alternatives, showing that (i) \ac{LLM}-generated images combined with textures are powerful representations for natural images, and (ii) scaling up the number of generative image procedures (by combining the procedures in Shaders-21k with those sampled from \acp{LLM}) improves overall performance, as predicted in \cite{baradad2022procedural}. We also note that the models trained with or without GPT-4 images achieve roughly the same performance. We end this section with the conclusion that \acp{LLM}, processing only textual inputs and outputs, can produce images with useful visual properties that complement previous procedural generation approaches for training visual systems.

\section{Conclusions}
We show that \acp{LLM} learn visual properties of the real world, and can depict them in the form of procedural image code. We do so by first analyzing the properties of the samples they generate, and showing that the model's competence %
can be further improved through text feedback from the model itself. 
Finally, we show that the images produced by these models can be used to train useful vision backbones for downstream tasks on natural images, gracefully complementing alternative approaches.

\newpage
\section*{Acknowledgement}
This research was supported by a grant from the MIT-IBM Watson AI lab. Adrián Rodríguez-Muñoz was supported by the LaCaixa Fellowship, Tamar Rott Shaham was supported by the Zuckerman STEM Leadership Program and the Viterbi Fellowship.

{
    \small
    \bibliographystyle{ieeenat_fullname}
    \bibliography{main}
}

\end{document}